\documentclass[sigconf]{acmart}
\AtBeginDocument{%
  \providecommand\BibTeX{{%
    \normalfont B\kern-0.5em{\scshape i\kern-0.25em b}\kern-0.8em\TeX}}}

\setcopyright{acmcopyright}
\copyrightyear{2018}
\acmYear{2018}
\acmDOI{XXXXXXX.XXXXXXX}

\acmConference[Conference acronym 'XX]{Make sure to enter the correct
  conference title from your rights confirmation emai}{June 03--05,
  2018}{Woodstock, NY}
\acmPrice{15.00}
\acmISBN{978-1-4503-XXXX-X/18/06}

\newcommand\given[1][]{\:#1\vert\:}
\newcommand\norm[1]{\left\lVert#1\right\rVert}
\acmConference[Conference acronym 'XX]{Make sure to enter the correct
  conference title from your rights confirmation emai}{June 03--05,
  2018}{Woodstock, NY}

\begin{document}

\newcommand{\xuyi}[1] {\textcolor{magenta}{#1}}
\newcommand{\etal}{\textit{et al}.}

%%
%% The "title" command has an optional parameter,
%% allowing the author to define a "short title" to be used in page headers.
\title{Relit-NeuLF: Efficient Relighting and Novel View Synthesis via Neural 4D Light Field}

\author{Zhong Li$^1$, Liangchen Song$^2$, Zhang Chen$^{1}$, Xiangyu Du$^{1}$, Lele Chen$^1$, Junsong Yuan$^{2}$, Yi Xu$^1$
}
\affiliation{%
  \institution{$^1$OPPO US Research Center\quad $^2$The State University of New York at Buffalo\quad \\
  \{zhong.li,zhang.chen,xiangyu.du,lele.chen,yi.xu\}@oppo.com}
  \country{}
  \ \textcolor{blue}{\url{https://oppo-us-research.github.io/RelitNeuLF-website/}} }
% \email{}

%%
%% By default, the full list of authors will be used in the page
%% headers. Often, this list is too long, and will overlap
%% other information printed in the page headers. This command allows
%% the author to define a more concise list
%% of authors' names for this purpose.
\renewcommand{\shortauthors}{Zhong Li et al.}

%%
%% The abstract is a short summary of the work to be presented in the
%% article.
\begin{abstract}
  In this paper, we address the problem of simultaneous relighting and novel view synthesis of a complex scene from multi-view images with a limited number of light sources. We propose an analysis-synthesis approach called Relit-NeuLF. Following the recent neural 4D light field network (NeuLF)~\cite{li2022neulf}, Relit-NeuLF first leverages a two-plane light field representation to parameterize each ray in a 4D coordinate system, enabling efficient learning and inference. Then, 
  %with coupled DecomposeNet and RenderNet, 
  we recover the spatially-varying bidirectional reflectance distribution function (SVBRDF) of a 3D scene in a self-supervised manner. A DecomposeNet learns to map each ray to its SVBRDF components: albedo, normal, and roughness. Based on the decomposed BRDF components and conditioning light directions, a RenderNet learns to synthesize the color of the ray. To self-supervise the SVBRDF decomposition, we encourage the predicted ray color to be close to the physically-based rendering result using the microfacet model. Comprehensive experiments demonstrate that the proposed method is efficient and effective on both synthetic data and real-world human face data, and outperforms the state-of-the-art results. We publicly released our code on GitHub. You can find it here: \textcolor{pink}{\url{https://github.com/oppo-us-research/RelitNeuLF}}
\end{abstract}

%%
%% The code below is generated by the tool at http://dl.acm.org/ccs.cfm.
%% Please copy and paste the code instead of the example below.
%%


%%
%% Keywords. The author(s) should pick words that accurately describe
%% the work being presented. Separate the keywords with commas.

%% A "teaser" image appears between the author and affiliation
%% information and the body of the document, and typically spans the
%% page.
\begin{teaserfigure}
  \includegraphics[width=0.9\textwidth]{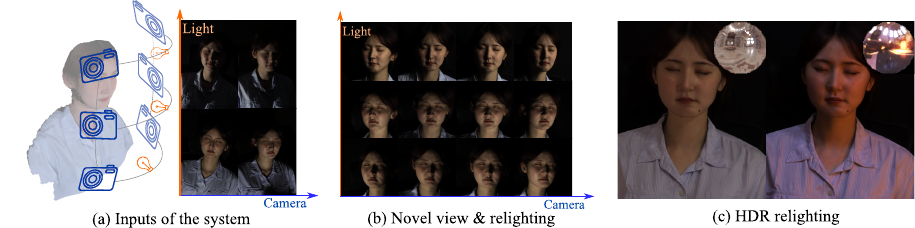}
  \caption{The proposed method takes (a) sparsely sampled multi-view images under different directional lights as input. After modeling the scene with the proposed Relit-NeuLF, our system can perform simultaneous novel view synthesis and relight with both (b) novel directional lights and (c) HDR light probes.}
  \label{fig:teaser}
\end{teaserfigure}

\maketitle

% \received{20 February 2007}
% \received[revised]{12 March 2009}
% \received[accepted]{5 June 2009}
\section{Introduction}
Novel view synthesis of real-world objects under different lighting conditions has been an important research topic for computer vision and computer graphics. A system that can perform simultaneous relighting and free-viewpoint rendering, thereby integrating a real-world object into a digital environment, has numerous commercial applications, especially when combined with real-time light estimation techniques~\cite{liu2022real,garon2019fast}. %in the visual effects industry and consumer electronics.

Traditional methods~\cite{kanade2007virtualized,collet2015high,zhang2017light,li2017robust,li20184d}, solve the free-viewpoint rendering problem based on the geometric reconstruction pipeline, which assumes the objects are mostly Lambertian. Since these methods do not fully model the complex surface reflection properties, the rendered results may be unsatisfactory. Taking human faces as an example, these methods do not model the spatially varying reflectance characteristics, such as diffuse and specular reflections.

To overcome these limitations, Light Stage~\cite{debevec2000acquiring} are used to capture the reflectance field of objects. With synchronized LED lights and cameras, a ``One-Light-at-A-Time'' (OLAT) image set can be obtained. Then, an input environment map can be decomposed into a linear combination of these OLAT images to relight the object in a photorealistic manner. This approach has a profound impact on the visual effects industry and inspires numerous follow-up works. To accurately capture the reflectance field, many light sources are required. However, due to physical constraints, only about one hundred light sources are usually installed on a light stage. Therefore, the system cannot produce sharp shadows and high-frequency details due to under-sampling. To solve this problem, Sun \etal ~\cite{sun2020light} propose a learning-based method to perform high-precision resampling of the under-sampled OLAT image set. However, simultaneous novel view synthesis during relighting remains unsolved.

%Sun et.al~\cite{sun2020light} propose a learning-based method to perform high-precision resampling of the undersampled OLAT image set to solve this problem, but it still cannot provide the free view synthesis while relighting the object. 

For image-based novel view synthesis, great progress has been made with the neural radiance field~\cite{mildenhall2020nerf} (NeRF), based on a volumetric representation. One of the limitations of NeRF is that the color and density of sample points along each ray must be integrated. Therefore, rendering with NeRF is time consuming. A few works ~\cite{hedman2021baking,yu2021plenoctrees,fridovich2022plenoxels,song2023nerfplayer} have attempted to accelerate NeRF by either using much more memory or a geometric \textit{prior}. While there has been substantial progress in novel view synthesis, relatively less effort has been spent in free-viewpoint rendering with relighting. 

In this paper, given sparsely sampled images illuminated by sparse LED light sources (as shown in Fig.~\ref{fig:teaser}), we present a two-stage analysis-synthesis MLP neural network for free-viewpoint rendering and relighting. Our model takes the camera ray direction and light direction as the input. We first analytically recover reliable normal, albedo, and roughness in a self-supervised manner (analysis stage), and then simulate the process of physically-based rendering to output the ray color (synthesis stage). 

Motivated by the recent work NeuLF~\cite{li2022neulf}, we use two-plane representation to parameterize a ray in space into a 4D coordinate. 
%The first analysis stage is a \textit{DecomposeNet}: the input ray goes through the DecomposeNet for SVBRDF decomposition and the outputs are the surface normal, albedo, and roughness. 
The analysis stage involves a \textit{DecomposeNet}, whose inputs are rays and the outputs are the decomposed surface normal, albedo, and roughness. The synthesis stage is a \textit{RenderNet}. With the predicted SVBRDF components from the DecomposeNet and a light direction, RenderNet is trained to generate the color for the input ray. The entire network is trained end-to-end and is supervised with both photometric loss and an SVBRDF rendering loss (using a microfacet model, \textit{e.g.},~\cite{karis2013real}). The SVBRDF parameters are self-supervised. We found that the SVBRDF decomposition of a ray makes it easier for the network to learn the implicit rendering process. 
% Thus, the model can generalize better for ray directions that are unseen during training, and the rendered outputs are more in line with the physical characteristics of the scene. \xuyi{can we back this up?}

Compared to NeRF's volumetric representation~\cite{mildenhall2020nerf}, our method utilizes a 4D light field representation~\cite{li2022neulf}. By directly mapping a ray and a light direction to a color value, we achieve efficient inference and low memory consumption. While this approach limits rendering to front views, it suffices for most free-viewpoint relighting applications, such as remote 3D video calls with changing backgrounds. Our contributions are:

% \vspace{-1.5mm}
\begin{itemize}
  \item {We propose a two-stage analysis-synthesis MLP network for efficient relighting and novel view synthesis, which maps ray and light direction to a color value directly.} 

   \item{Our proposed DecomposeNet can recover the SVBRDF (surface normal, albedo, and roughness) of a 3D scene in a self-supervised manner using a set of camera viewpoints and a limited number of light sources as inputs.}
 
   \item {We created a multi-identity multi-view OLAT dataset of real human faces and a synthetic multi-view multi-lighting dataset. We plan to release our dataset to facilitate future research on relighting and novel view synthesis.}
  
\end{itemize}
\section{Related Works}

Our method simultaneously synthesizes novel views and upsamples the sparse light sources. These two problems subsume many computer vision and graphics-related tasks and have been key topics for decades. Below we briefly review the related work on light angular up-sampling and novel view synthesis, as well as recent efforts on these using neural rendering.

\textbf{Light angular upsampling:} Traditional methods upsample discrete light source input using geometric modeling, a technique distinct from light stage data upsampling~\cite{debevec2000acquiring}. Such methods undertake geometric and photometric stereo reconstruction of objects, revealing albedo, normal, and specular normal~\cite{woodham1980photometric, ma2007rapid,xu2010high}, and subsequently employ predefined BRDFs for different lighting conditions. Tunwattanapong \etal employ parametric models~\cite{tunwattanapong2011practical}. Deep learning adaptations, distinct from traditional light stage data, have been applied for single-image human relighting~\cite{sun2019single,pandey2021total,kanamori2018relighting}. Shu \etal presented a human face illumination transfer, with some limitations in compatibility~\cite{shu2017portrait}. Challenges persist with predefined BRDFs, especially when interpreting complex facial features. Methods from both Masselus \etal~\cite{masselus2004smooth} and Rainer \etal~\cite{rainer2019neural, rainer2022neural} seek to address these challenges, but results may lack detail. Sengupta \etal highlight challenges in expressing objects with complex materials, pointing out issues with predefined analytic BRDFs~\cite{sengupta2018sfsnet}. Xu \etal introduced a neural network for image-based relighting, although shadow accuracy remains problematic~\cite{xu2018deep}. Our approach accentuates shadow accuracy and includes novel view synthesis. Lastly, Sun \etal~\cite{sun2020light} provided a method for relighting faces, albeit without supporting novel view synthesis.

\begin{figure*}[!htb]
    \centering
    \includegraphics[width=0.9\textwidth]{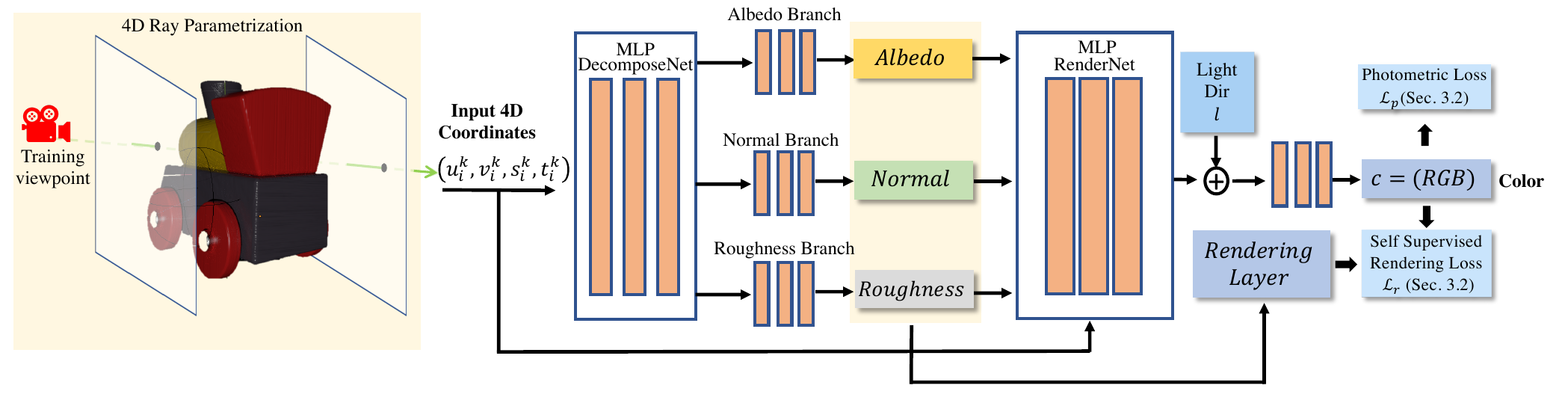}
    \caption{An overview of our proposed Relit-NeuLF. The input is the 4D coordinate of a ray and a light direction. The output is the RGB radiance of the ray under the light direction. Our DecomposeNet first takes the 4D coordinate as input and outputs SVBRDF parameters. Next, the SVBRDF parameters together with the light direction are fed into an implicit rendering network (RenderNet) to synthesize the target color. The network is trained end-to-end with photometric loss and self-supervised rendering loss.}
    
    % By optimizing the differences between the predicted colors and ground-truth colors, NeLF can faithfully learn the mapping between a 4D coordinate that characterizes the ray and its color. We also build a depth branch to let the network learn the per ray scene depth by self-supervised losses $\mathcal{L}_{mp}$ and $\mathcal{L}_d$.}
    % % {We also enforce the frequency distribution similarity (the Fourier Sparsity Loss $\mathcal{L}_s$, see Sec.~\ref{sec:fourier}) and neighboring ray smoothness (the Ray Bundle Loss $\mathcal{L}_r$, see Sec.~\ref{sec:raybundle}) to avoid overfitting and to solve aliasing problems, respectively.}}
    \label{fig:pipeline}
\end{figure*}

\textbf{Novel view synthesis:} For generating novel views from a set of images, traditionally, multi-view stereo algorithms ~\cite{furukawa2009accurate,zhang2017light} can be used to reconstruct a textured mesh. To achieve higher photo realism, this approach requires more vertices and high-resolution texture maps. It does not model view-dependent effects either. Image-based rendering, such as light field~\cite{levoy1996light} and lummigraph~\cite{gortler1996lumigraph}, can render view-dependent effects by reducing the 5D radiance field in space to a 4D light field, but the quality of the novel view synthesis is affected by the sampling density of the input ~\cite{chai2000plenoptic}. Subsequent works use geometry proxy for better view interpolation~\cite{unstructuredlumigraph, wood2000surface}, but the results also depend on the accuracy of the proxy and are sensitive to occlusions. Recently, deep learning has been applied to generate novel views from sparse images. Neural networks are typically used to generate an implicit intermediate representation, and a rendering algorithm is developed based on the representation ~\cite{thies2019deferred,zhou2018stereo,sitzmann2019deepvoxels,sitzmann2019scene,mildenhall2019local}. A recent seminal work NeRF~\cite{mildenhall2020nerf,verbin2022ref} generate unprecedented results. It uses an MLP network to regress 5D coordinates (location in space and view direction) to the view-dependent radiance and density volume. Nevertheless, it requires aggregating the samples along each ray during rendering, which is time-consuming. Although there are recent solutions to accelerate NeRF \cite{hedman2021baking,fridovich2022plenoxels,yu2021plenoctrees}, they typically need to consume more memory. On the other hand, neural light field approaches \cite{li2022neulf,sitzmann2021lfns,suhail2022light,attal2022learning,sun2021nelf,song2023harnessing} use a neural network to directly establish the mapping between ray direction and color; therefore, no complicated ray marching is needed. These methods improve the rendering speed while maintaining low memory consumption. But none of these methods model environment lighting and thus cannot relight the scene under unseen illumination.

~\textbf{Simultaneous free-viewpoint rendering and relighting:} To render novel views with relighting effects simultaneously, an effective method is to sample sufficient reflection fields from multiple viewpoints and under multi-directional point lights and then upsample across viewpoints and illumination. A pioneer work of Debevec \etal~\cite{debevec2000acquiring} built a light stage to collect One-Light-at-A-Time (OLAT) data, and then linearly combine reflectance fields to relight the subject. Although the light stage can collect reflectance fields from hundreds of light directions, the sampling rate is still limited. It can only be applied to low-resolution environment maps for re-lighting, so high-frequency lighting effects are not supported. Moreover, the light stage can only perform relighting in a few fixed viewing angles. 

A recent work models texture maps to achieve relighting effects with known subject geometry~\cite{zhang2021neural}. Zhang \etal and Boss \etal propose methods to decompose the shape and reflectance of the subject to achieve new perspective rendering and relighting without known lighting conditions ~\cite{zhang2021nerfactor,boss2021neural,boss2021nerd,boss2020two,zhang2022modeling,zhang2021physg,boss2022samurai,munkberg2022extracting,yao2022neilf,zhang2022iron}. However, both methods can only leverage a relatively low-resolution environment lighting map, hence the results have no high-frequency details. NeRV~\cite{srinivasan2021nerv} and PS-NeRF~\cite{yang2022ps} improve on NeRF and decomposes shape and reflectance in the case of known multi-illumination. Our method is based on the light field representation ~\cite{li2022neulf} and proposes a two-step network for simultaneous novel view synthesis and relighting. With internet image collections, Zhang et al.~\cite{zhang2021ners} reconstruct both the geometry and material appearance of the object, however, it can only model the genus-zero topology.
%We use the light field representation as the input, the first step neural nets self-supervised decomposed light BRDF components as the next level network input, and train the second step implicit rendering network to get the ray color in different lighting directions. Therefore 
Our method does not require a geometry proxy for high-frequency relighting and achieves fast free-viewpoint rendering with less memory consumption.

\section{Proposed Method}

In this paper, we present a simple yet efficient framework that can generate novel views under novel directional, given sparse multi-view One-Light-at-A-Time(OLAT) images as input.
%captured in a light stage~\cite{debevec2000acquiring}. 
To achieve this, we extend a neural 4D light field representation NeuLF~\cite{li2022neulf} (Section~\ref{Sec:NeuLF}) with relighting capability by using SVBRDF decomposition. Given the set of (ray, color) pairs from known views, to guide the network with physically based rendering, we first design a DecomposeNet that disentangles each ray's intrinsic SVBRDF components (albedo, normal, and roughness). 
We then use a RenderNet to use albedo, normal, roughness, and ray direction as input with light direction as conditioning to regress ray color (Section~\ref{Sec:Relit-NeuLF}). Fig.~\ref{fig:pipeline} shows our pipeline.

\subsection{NeuLF Overview}
\label{Sec:NeuLF}
We review the NeuLF representation in this section. As the radiance along a ray is constant if viewing from outside the convex hull of the scene, a 3D scene can be represented as a 4D light field using two plane parameterization~\cite{levoy1996light}. Each ray $r$ from a camera with known camera poses will intersect with both the $uv$ and the $st$ plane and can be uniquely represented using the coordinate of the intersections $r = \{u,v,s,t\}$.

% \begin{figure}
%     \centering
%     \includegraphics[width=\linewidth]{Figure/2pp.pdf}
%     \caption{Two-plane representation of the light field. Each ray is parametrized by a 4D coordinate $u,v,s,t$, which are the intersects on the $uv$ and $st$ plane.}
%     \label{fig:2pp}
% \end{figure}

NeuLF~\cite{li2022neulf} formulates a mapping function $f_c$ with a multilayer perceptron (MLP). Given a set of $N$ calibrated input images $\{I_1,I_2,...,I_n\}$, it extracts the 4D ray parametrization of each pixel and its corresponding color: $\left(u_i^k,v_i^k,s_i^k,t_i^k\right) \rightarrow c_i^k, (k=1...N, i=1...N_k)$, where $c_i^k$ is the color of the $i$-th pixel in the $k$-th image. $N_k$ is the total number of pixels in the $k$-th image. The goal of the network $f_c$ is to learn the mapping from the 4D coordinate of a ray to the corresponding color value. The MLP network parameters $\Theta$ can be learned by minimizing the following photometric loss $\mathcal{L}_p$:

\begin{equation}
    \label{eqn:photometricloss}
    \mathcal{L}_p = \sum_{k=1}^{M}\sum_{i=1}^{N_k}\norm{f_c\left(u_i^k,v_i^k,s_i^k,t_i^k \given \Theta\right)-c_i^k}_2.
\end{equation}

To render a novel view, NeuLF simply queries the network for each ray in the novel view.

% \vspace{-2mm}
\subsection{Relit-NeuLF}
\label{Sec:Relit-NeuLF}
The NeuLF representation does not consider the changes of incident light direction. In other words, NeuLF is only able to represent a scene under a fixed baked-in illumination. Given a set of OLAT input images $\{I_i^l,i=1,...,N, l=1,...,L\}$, we have $N$ different viewpoints and each viewpoint is illuminated by $L$ different directional lights. In total, we have $N \times L$ images. To perform lighting interpolation and novel view synthesis simultaneously, one straightforward approach is to extend NeuLF to take lighting direction as input. Instead of only mapping 4D coordinate to color value as in Eq.~\ref{eqn:photometricloss}, we can train a mapping $f_c^l((u_k,v_k,s_k,t_k),l) \rightarrow c_k^l$ for each input view $i$.
% that maps input 4D coordinates together with a light direction to a color value. For each input view $i$, we train a mapping $f_c^l((u_k,v_k,s_k,t_k),l) \rightarrow c_k^l$. 
With this, we can simulate the light transport (LT) of a given 4D coordinate ray,\textit{ i.e.}, how color value changes with varying lighting direction.

% \xuyi{***this part is not very clear, do we have one mapping per image or per lighting? Text says per image however notation says per light***}

However, this vanilla method simply regresses a light-conditioned ray to a color value. Without implicitly modeling the scene representation, synthesis results obtained by such a method lacks high-frequency detail and are unable to recover complex reflection and refraction effects (see Sec.~\ref{sec:svbrdfd} for details). We hence propose a two-stage network designed with principles inspired by physically-based rendering techniques. In the first stage, we create a DecomposeNet that analyzes a ray $r$'s SVBRDF components, including normal, albedo, and roughness. Then, these SVBRDF components are fed into a second network named RenderNet to learn the inverse rendering function. The SVBRDF components can be recovered without ground truth by self-supervision. We will describe each network in detail.

% \begin{figure}
%     \centering
%     \includegraphics[width=\linewidth]{Figure/sampleInput.pdf}
%     \caption{Real data captured by our multi-camera \& multi-light system. Our system is capable of capturing OLAT images from 13 camera views and 40 light directions.}
%     \label{fig:input}
% \end{figure}
% How about the rend
~\textbf{DecomposeNet:} Given the input $N \times L$ images (Fig. \ref{fig:hardware} shows sample images), we build a MLP network that takes 4D coordinate $r=\{u,v,s,t\}$ for each pixel as input, and outputs SVBRDF parameters: surface normal ($N$), diffuse albedo ($A$) and specular roughness ($R$). Considering that the SVBRDF components are closely correlated, the DecomposeNet first extracts a shared feature and then uses three decoders for three different components. Compared with using a completely separate branch for each SVBRDF component, this shared structure improves efficiency and reduces the risk of over-fitting. It is also worth nothing that our method can recover SVBRDF components in a self-supervised fashion without knowing the ground truth. We will illustrate our self-supervised loss later in the following paragraph. The network DecomposeNet$(\cdot)$ is represented as: 
\begin{equation}
    \label{eqn:DecomposeNet}
    N,A,R = DecomposeNet(r \given \Theta_d),
\end{equation}
where $r \in \{I_i^l,i=1,...,N, l=1,...,L\}$.\\
% \subsection{RenderNet}
% \label{Sec:RenderNet}

~\textbf{RenderNet:} Prior works leverage predefined SVBRDF components to analytically render the scene~\cite{li2018learning}. However, using predefined SVBRDF (\textit{e.g.}, microfacet renderer \cite{karis2013real}) is not a universal rendering solution. It cannot model complex objects and materials at the same level of fidelity as real photos, such as human faces' subsurface scattering and specularity distribution. Failing to model them will compromise the photorealism of the rendering results. We propose to train a network to implicitly model the real-world rendering process, using the microfacet renderer and ground truth for weak supervision. The RenderNet takes the estimated SVBRDF components (albedo ($A$), normal ($N$), and roughness ($R$)) as inputs. 
%Let RenderNet$(\cdot)$ be the network function we would like to train. 
To render a ray under specific light direction \textit{l}, we train a fully connected network:

\begin{equation}
    \label{eqn:RenderNet}
    c_{pred} = RenderNet(N,A,R,r,l \given \Theta_r ),
\end{equation}
where $r \in \{I_i^l,i=1,...,N, l=1,...,L\}$.

\begin{figure}
    \centering
    \includegraphics[width=0.9\linewidth]{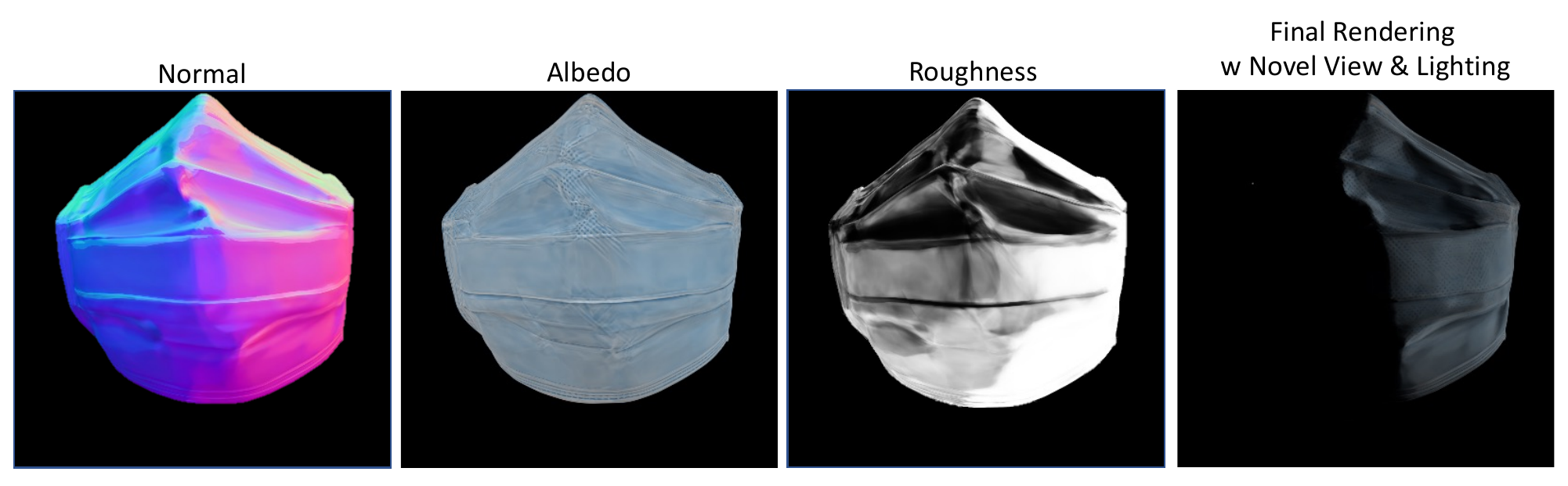}
    \caption{Our SVBRDF decomposition results. From left to right: surface normal map, surface albedo, surface roughness, and rendering results under novel viewpoint and light direction.}
    \label{fig:Depth_fig}
\end{figure}

\textbf{Loss function:}
We train both \textbf{DecomposeNet} and \textbf{RenderNet} in an end-to-end manner. For each ray in camera view $i$ and light direction $l$, we can compute the predicted color $c_{pred}$ as follows:

\begin{equation}
    % \label{eqn:Loss function}
    c_{pred} = RenderNet(DecomposeNet(r \given \Theta_d),r,l \given \Theta_r).
    % \mathcal{L}_p = \sum_{k=1}^{M}\sum_{i=1}^{N_k}\norm{DecomposeNet(r \given \Theta_d) \given \Theta)-c_i^k}_2
\end{equation}

During training, we leverage two data loss terms $\mathcal{L}_p$ and $\mathcal{L}_m$, and one normalization term $\mathcal{L}_n$. We use $\mathcal{L}_p$ to minimize the multi-view photometric loss. To self-supervise the SVBRDF decomposition, we use $\mathcal{L}_m$ to enforce the network output to be close to the physically-based microfacet rendering result. We choose the microfacet model~\cite{karis2013real} instead of the Phong model because the microfacet model can produce more realistic rendering results.
% , because the microfacet model has been proven to exhibit more realistic rendering results on refraction and reflection rough surfaces, which is common in our real world. 
To regularize the normal vector to be physically plausible, we use $\mathcal{L}_n$ to encourage the normal $N$ to have a norm of $1$. 

\begin{equation}
    % \label{eqn:Loss function}
    % C_{pred} = RenderNet(DecomposeNet(r \given \Theta_d),r,l | \Theta_d)
    \mathcal{L}_p = \norm{c_{pred}-c}_2,
\end{equation}

\begin{equation}
    % \label{eqn:Loss function}
    % C_{pred} = RenderNet(DecomposeNet(r \given \Theta_d),r,l | \Theta_d)
    \mathcal{L}_m = \norm{c_{pred}-M(N,A,R,v,l)}_2,
\end{equation}

\begin{equation}
    % \label{eqn:Loss function}
    % C_{pred} = RenderNet(DecomposeNet(r \given \Theta_d),r,l | \Theta_d)
    \mathcal{L}_n = \norm{1 - N^TN}_2,
\end{equation}
where $M$ is the microfacet BRDF rendering model~\cite{karis2013real}, $A$ is the diffuse albedo, and $v,l$ denote the view direction and light direction respectively. Let $h=\frac{v+l}{\|v+l\|}$ be the half vector, then we have

\begin{equation}
    \label{eqn:Loss function}
    % C_{pred} = RenderNet(DecomposeNet(r \given \Theta_d),r,l | \Theta_d)
    M(N,A,R,v,l) = \frac{A}{\pi} + \frac{D(h,N,R)F(v,h)G(l,v,N,R)}{4(N \cdot{l} )(N \cdot{v} )},
    % \mathcal{L}_m = \norm{c_{pred}-M(N,A,R)}_2
\end{equation}
where $D(h,N,R) = \frac{a^2}{\pi((N\cdot{h})^2(a^2-1)+1)^2}$ is normal distribution function (NDF), $F(v,h)=F_0 + (1-F_0)2^{(-5.55473(v\cdot{h})-6.98316)(v\cdot{h})}$ is Fresnel term~\cite{lvovsky2013fresnel}, and $G(l,v,N,R)=\frac{N\cdot{v}}{(N\cdot{v})(1-k)+k}\cdot{\frac{N\cdot{l}}{(N\cdot{l})(1-k)+k}}$ is geometry term. The extra variables in the equations are $F_0=0.05$, $a=R^2$, and $k=\frac {(R+1)^2}{8}$.

Fig.~\ref{fig:Depth_fig} shows an example of our decomposition and rendering results. It can be seen that \textbf{DecomposeNet} successfully decomposes the BRDF components with the self-supervision of the microfacet model, and can generate rendering results under unseen light direction.

% \textcolor{red}{regularization term for normal:todo}
% rendering time:////

% \subsection{Generating the training data}
\subsection{Implementing Details}
We train the networks using a combination of the photometric loss, microfacet rendering loss, and normal regularization loss:

% \lc{we do not need use mathcal L}
\begin{equation}
    % \label{eqn:Loss function}
    % C_{pred} = RenderNet(DecomposeNet(r \given \Theta_d),r,l | \Theta_d)
    \mathcal{L} = \lambda_m\mathcal{L}_m + \lambda_p\mathcal{L}_p  + \lambda_n\mathcal{L}_n.
\end{equation}

The weight coefficients for the three losses are $\lambda_m=0.1$, $\lambda_p=1$ and $\lambda_n=0.01$. 
We conduct experiments on both synthetic data and real captured data. For real data, to estimate camera poses, we adopt ~\cite{zhang2000flexible} and use the planar pattern to calibrate the intrinsic and extrinsic parameters of each camera. To calibrate the directions of light sources, we capture images of a chrome ball under all the lighting conditions and use the brightest spot to identify the direction of each light.

We normalize the input 4D coordinate ($u,v,s,t$) to the range $[-1,1]$, and then go through the 8 layers of fully connected ReLU in DecomposeNet to obtain the 256-dimensional shared feature vector. Then the feature vector is split into three SVBRDF parameter branches: $N$ (Normal), $A$ (Albedo), $R$ (Roughness), each of which contains 1 fully-connected layer. The Normal and Albedo branches have 3-dimensional outputs and use ReLU activation, while the Roughness branch has a 1-dimensional output and uses sigmoid activation. 
Then we concatenate $N$, $A$, $R$ with ray coordinate $\{u,v,s,t\}$ as the input to the RenderNet, which uses 8 layers of fully connected and ReLU to encode its input into a 128-dimensional feature vector. The feature vector is concatenated with 3-dimensional light direction $l$, then goes through two fully-connected layers with sigmoid activation, and finally outputs 3-channel RGB radiance. In terms of training details, we set the batch size of each iteration to 8192 and use Adam optimizer~\cite{da2014method}. The initial learning rate is set to $3 \times 10^{-4}$ and decays 0.995 per epoch. In the real data experiment, we use an input of $13$ camera views with $500 \times 750$ resolution and $40$ LED light sources. The training takes 5 hours to complete on 1 NVIDIA RTX 3090 graphics card, while the inference takes 150ms per frame under the same resolution.

\section{Experiment}
%-------------------------------------------------------------------------
To demonstrate the effectiveness of the proposed framework for high-quality relighting and novel view synthesis, we conduct experiments on both rendered synthetic data and captured real-world datasets. We first discuss the acquisition of real data and synthetic data in Section~\ref{sec:data_setup}. 
% Then we demonstrate the effectiveness of the relighting and view synthesis results respectively. 
Then we show both quantitative and qualitative results of novel view synthesis and relighting in Section~\ref{sec:results}. 
\textcolor{black}{We report rendering speed and quality with comparison to an extended NeRF-based method in Section~\ref{sec:speed}.} 
We also show our SVBRDF decomposition results and perform an ablation study to illustrate its importance in Section~\ref{sec:svbrdfd}.  We refer readers to our supplemental video, which showcases our results on dynamic relighting and view synthesis.

\begin{figure}[!h]
    \centering
    \includegraphics[width=0.9\linewidth]{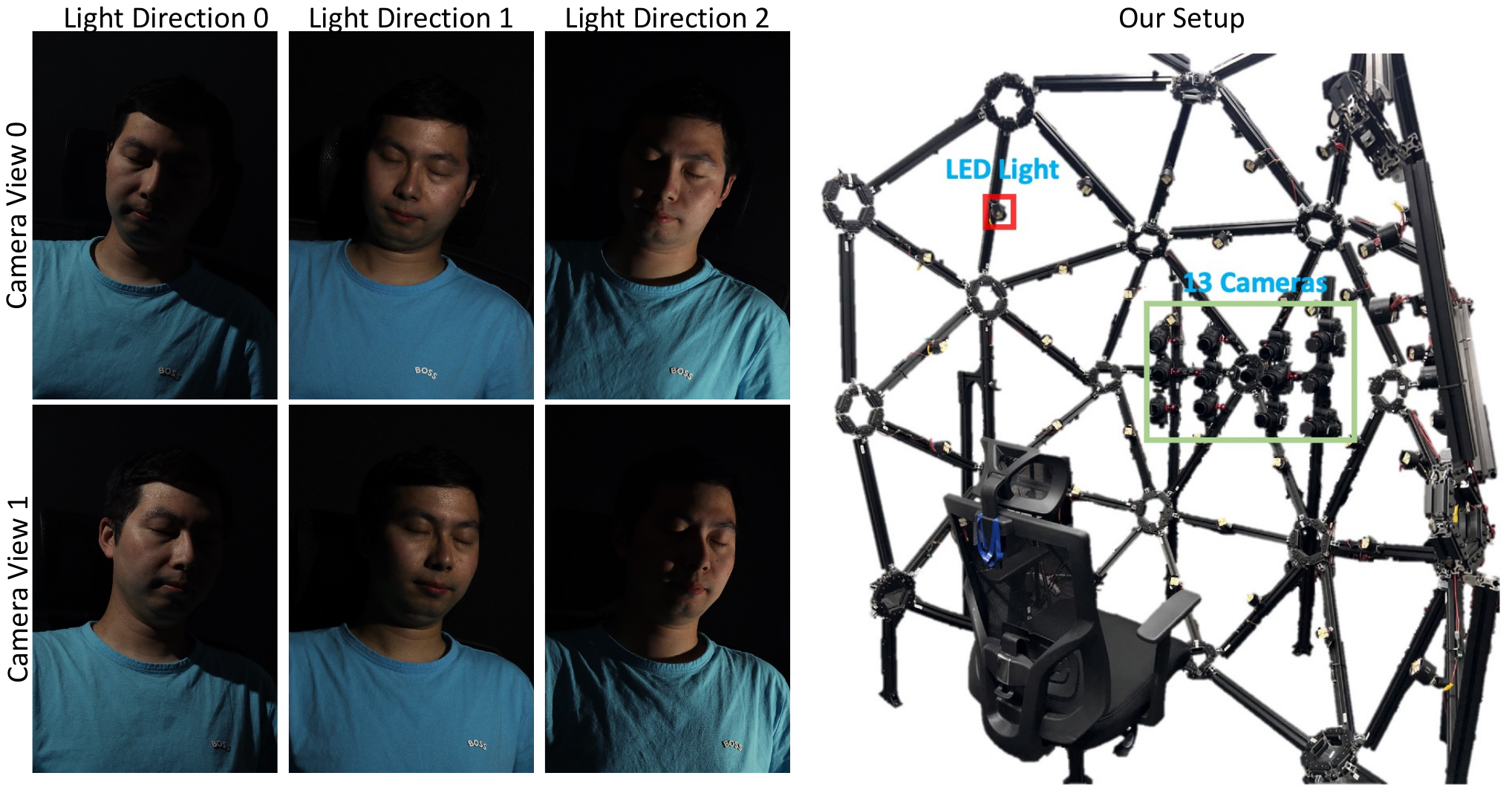}
    \caption{Our multi-camera \& multi-light capture system, with sample captured images. The system contains 13 synchronized DSLR cameras and 40 LED lights. We use an Arduino microcontroller to control the cameras and lights.}
    \label{fig:hardware}
\end{figure}

\begin{figure*}[!h]
    \centering
    \includegraphics[width=0.9\linewidth]{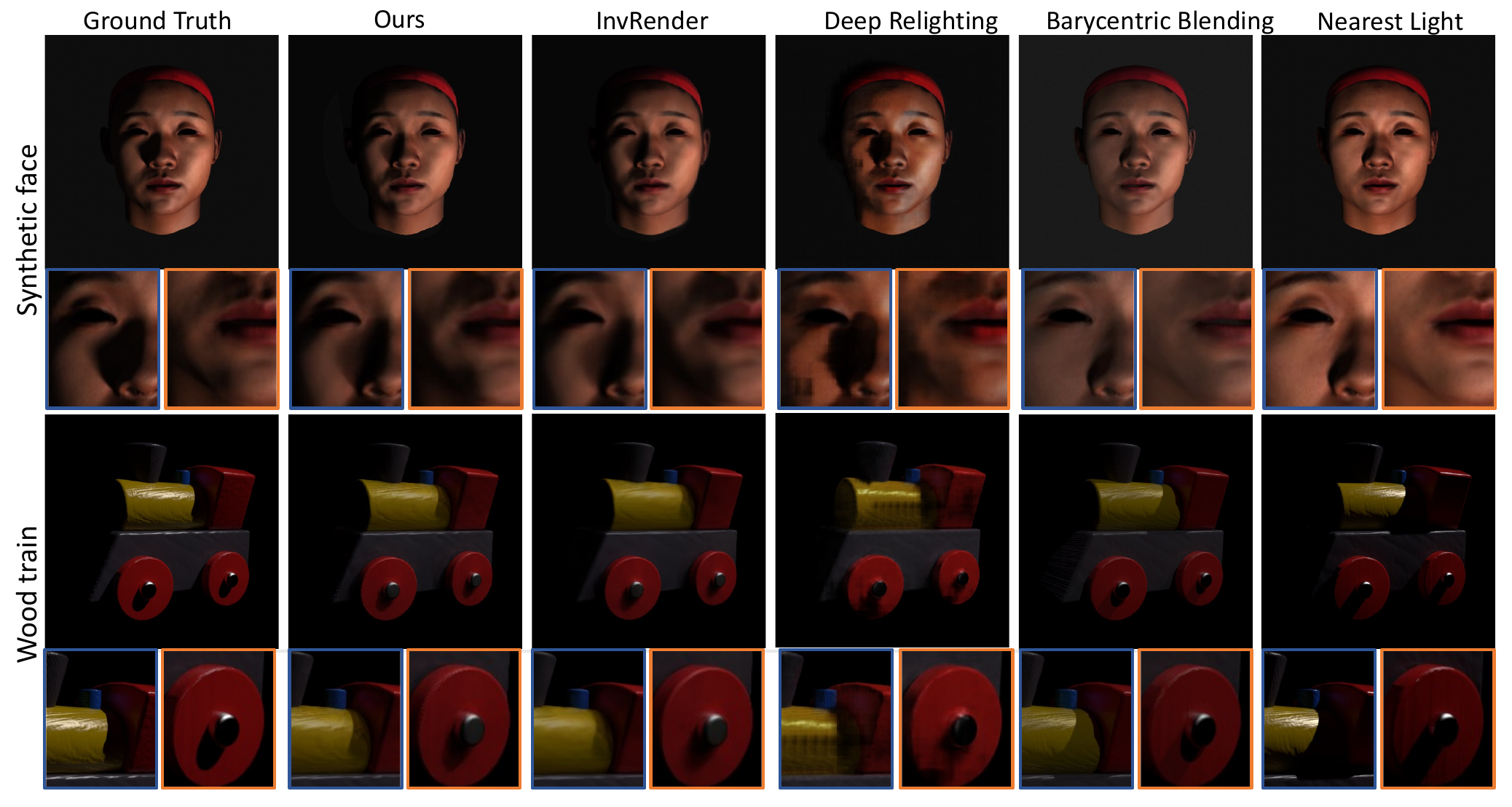}
    \caption{Qualitative results of relighting with an unseen light direction under fixed viewpoints. From left to right: Ground truth, our method, InvRender~\cite{zhang2022modeling}, Deep Relighting~\cite{xu2018deep}, barycentric blending, and nearest light image.}
    \label{fig:QuantiRes}
\end{figure*}

\subsection{Real and Synthetic Data Setup}\label{sec:data_setup}
~\textbf{Real data acquisition} Obtaining multi-view images under different lighting directions is not an easy task. To the best of our knowledge, there is no such dataset publicly available. Similar to ~\cite{guo2019relightables}, we built a hemisphere multi-view OLAT (one light at a time) image acquisition system. As shown in Fig.~\ref{fig:hardware}, our system consists of 13 synchronized Canon 850D DSLR cameras and 40 LED light sources. We have shown 4 subjects in our experiments. For each subject, we acquire $13 \times 40$ OLAT images. These OLAT images sparsely collected the reflectance data on the front hemisphere of the subject. 
% An example of the OLAT images is shown in Fig. \ref{fig:input}.

During data collection, we require all subjects to remain approximately still. It takes 5 seconds to complete the OLAT acquisition of 13 viewing angles and 40 LED light sources. 
% Despite the short time slot, it is difficult to keep the subject still, so we ask the subject to rest his head on the headrest of the chair, 
From the collected data of each subject, we randomly selected 1 out of 13 views and 3 out of 40 light conditions for testing, and use the rest for model training.

\begin{figure}[!h]
    \centering
    \includegraphics[width=0.9\linewidth]{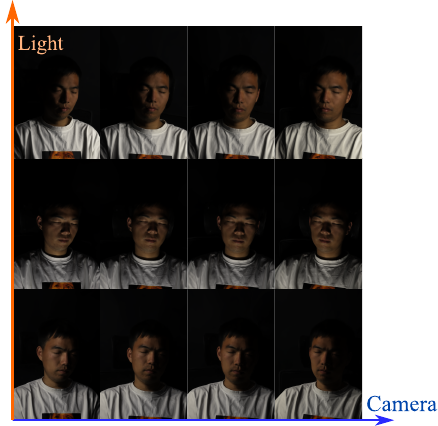}
    \caption{Qualitative results on simultaneous novel view synthesis and relighting. We show results on real data. The x-axis is different novel viewpoints while the y-axis is different lighting directions}
    \label{fig:fvvrelight}
\end{figure}

~\textbf{Synthetic data rendering} Additionally, we used Blender's physically based path tracer renderer and rendered 3 textured objects: synthetic face, wood train, and face mask. We set up $5 \times 5$ camera views on the front hemisphere, set $105$ directional light sources around the full sphere, and render at a resolution of $800 \times 800$ pixels. Each camera differs by 10 degrees and each light source differs by 25 degrees on the sphere. We randomly select 2 out of 25 views and 6 out of 105 lights for testing, and the rest for model training.

\addtolength{\tabcolsep}{-2.5pt}    
\begin{table*}[!h]
\centering
\caption{Metrics on relighting (averaged across test lights) for each scene in the Synthetic/Real dataset.\label{tab:quantitative}}
\resizebox{\textwidth}{!}{%
\begin{tabular}{l|ccccc|ccccc|ccccc}
\hline
% &       &    PSNR $\uparrow$     &       & & & SSIM $\uparrow$ &    && &LPIPS $\downarrow$  & \\ %\hline
& \multicolumn{5}{c|}{PSNR$\uparrow$} & \multicolumn{5}{c|}{SSIM$\uparrow$} & \multicolumn{5}{c}{LPIPS$\downarrow$} \\
% & NeRF~\cite{mildenhall2020nerf}  & LLFF~\cite{mildenhall2019local}  & SRN~\cite{srn2019}   & NSVF~\cite{Liu20neurips_sparse_nerf}  & Ours  & NeRF~\cite{mildenhall2020nerf}  & LLFF~\cite{mildenhall2019local}  & SRN~\cite{srn2019}   & NSVF~\cite{Liu20neurips_sparse_nerf}  & Ours         \\ \hline
& Nearest &  Barycentric  & Deep Relighting & InvRender  & Ours  & Nearest &  Barycentric  & Deep Relighting  & InvRender & Ours  & Nearest &  Barycentric  & Deep Relighting & InvRender & Ours\\ \hline

Toy train     & 32.79 & 33.60 & 33.43 &33.82 & \textbf{34.40} & 0.914 & 0.918 & 0.879 &0.919 & \textbf{0.930} & 0.078 & 0.082 & 0.084 &0.086 & \textbf{0.075}      \\
Face Cover     & 28.03 & 28.16 & 27.34 &27.96 & \textbf{28.79} & 0.793 & 0.801 & 0.763 &0.804 & \textbf{0.810} & 0.150 & 0.163 & 0.203 &0.160 & \textbf{0.130}      \\
Face         & 32.65 & 32.48 & 31.42 &32.88 & \textbf{33.14} & 0.820 & 0.808 & 0.754 &0.827 & \textbf{0.833} & 0.071 & 0.062 & 0.068 &0.081 & \textbf{0.075}     \\
Real     & 27.49 & 27.07 & 27.23 &27.86 & \textbf{28.33} & 0.732 & 0.745 & 0.708 &0.747 & \textbf{0.771} & 0.143 & 0.162 & 0.155 &0.141 & \textbf{0.132}      \\
\hline
% Real2       & 17.31 & \textbf{28.60} & 27.79 & 27.12 & 0.412 & \textbf{0.928} & 0.898 & 0.881 & 0.279 & \textbf{0.168} & 0.276 & 0.263      \\
% Crest           & 15.91 & \textbf{21.23} & 20.30 & 20.11 & 0.209 & \textbf{0.757} & 0.670 & 0.653 & 0.304 & \textbf{0.162} & 0.315 & 0.410      \\ \hline
\end{tabular}
}
\label{Tab:quantitative}
\end{table*}

\subsection{Relighting and View Synthesis Results}\label{sec:results}
~\textbf{Relighting} Our Relit-NeuLF model can generate rendering results under novel viewpoints and novel lighting directions.
As shown in Fig.~\ref{fig:QuantiRes}, we show qualitative relighting results for different synthetic data. 
We compared our method with two state-of-the-art solutions: InvRender~\cite{zhang2022modeling}, which generates novel view synthesis and relighting under unknown illumination, and Deep relighting~\cite{xu2018deep}, which presents an image-based relighting method using sparse predefined directional lights.
In addition, we implement two baseline solutions for comparison: barycentric blending and nearest light in the training set. 
Compared to the hold-out GT validation image, our results surpass the other methods in terms of shadow shape and clarity (\textit{e.g.}, the nose shadows in the zoom-in figures of the first row). 
Although the barycentric blending and nearest light methods can maintain a high-frequency appearance, their shadows have largely incorrect coverage.
This demonstrates that these methods, although widely adopted for relighting, have high requirements on the denseness of input lighting directions and are not suitable when the input is sparse.
Compared with Deep Relighting, our method produces more accurate shadow shapes, more natural specular highlights, and much fewer appearance artifacts, leading to more realistic results. While InvRender produces relatively good results, the outputs are slightly blurry and exhibit less accurate shadows, potentially due to the method's inability to generalize to single point light illuminations.
% ~\textcolor{red}{zoom in figure}
%  ~\textcolor{red}{Compare with Nerd or Nerd PIL}

For quantitative evaluation, we use three metrics: PSNR (Peak Signal-to-Noise Ratio, higher is better), SSIM (Structural Similarity Index Measure, higher is better), and LPIPS (Learned Perceptual Image Patch Similarity, lower is better). 
The first three rows of Table.~\ref{tab:quantitative} shows the metrics on the synthetic data. Our method consistently outperforms the other three with noticeable margins.
% , we evaluate these three metrics on synthetic and real experiments. We find that our results are better than other methods.
The last row compares the metrics on real captured human faces, where our method also has the highest performance.

\begin{figure}[!h]
    \centering
    \includegraphics[width=0.9\linewidth]{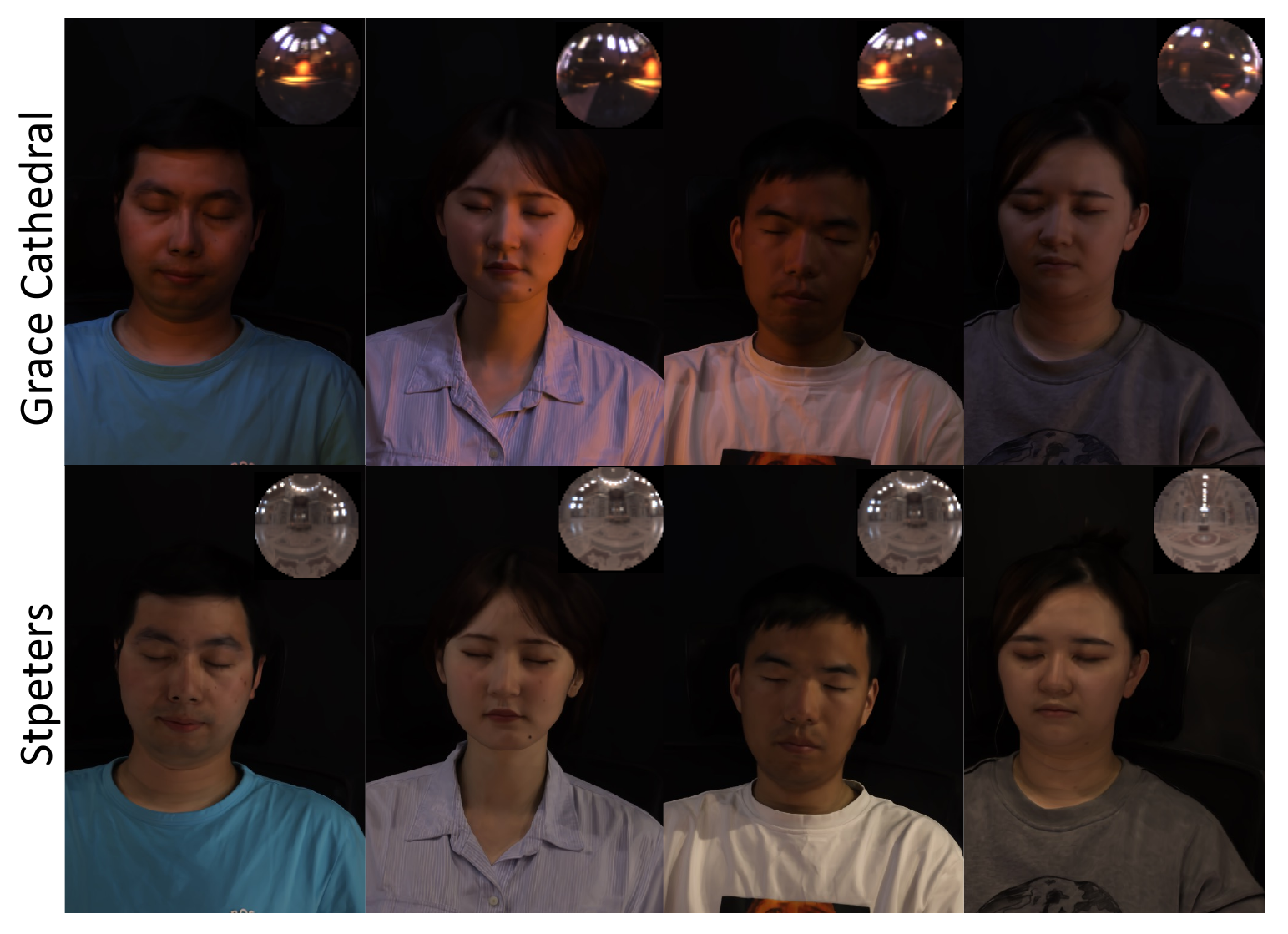}
    \caption{Our HDRI relighting results. Each subject is relit by two different environment maps. Note that the subject's appearance correctly reflects the dominant light source in the environment map. }
    % \textcolor{red}{how about more results?}}
    \label{fig:HDRlight}
\end{figure}

~\textbf{HDRI relighting} As shown in Fig.~\ref{fig:HDRlight}, we demonstrate the ability of our method to synthesize visually pleasing relighting under arbitrary HDRI environment maps. 
Because our method can recover the reflectance under novel lighting directions with a high angular resolution, we can relight the object by treating an HDRI environment map as a collection of OLAT lighting conditions. 
Unlike ordinary light stage which can only sample hundreds of OLAT images, we use our trained model to generate 3096 OLAT images, each corresponding to a different lighting direction. Then these OLAT images are linearly superimposed to produce the final relighting result, where the linear combination weights are computed based on the HDRI environment map.
We have shown four HDRI relighting results across different human identities using the light probe images from ~\cite{debevec2008rendering}. Note that for each row in Fig.~\ref{fig:HDRlight}, we rotate the light probe images by different angles. 
Our relighting results are physically natural, and the specular highlights on the faces correctly reflect the color and position of the highlights in the light probe images.
Meanwhile, the cast shadows are also visually plausible. 
Please see our supplemental video for more dynamic relighting results.

% ~\textcolor{red}{Compare with Nerd or Nerd PIL}

% thus we could clearly observe the dominant lighting effect on human face and casting plausible shadow direction

~\textbf{Simultaneous view synthesis and relighting} As shown in Fig.~\ref{fig:fvvrelight}, we qualitatively show our rendering results of two real human subjects under simultaneous novel view synthesis and relighting.
In the figure, the x-axis is different novel viewpoints while the y-axis is different lighting directions.
% We show our results on two subjects, where the abscissa is changing the viewpoints and the ordinate is the direction of changing the light source. 
The lighting effects (\textit{e.g.}, shadows and diffuse reflection) are consistent across different viewpoints. 
At the same time, high-frequency details (\textit{e.g.}, specular highlights on the foreheads) are successfully recovered and naturally move in accordance with viewpoint and lighting direction. 
We reason that this is because our RenderNet implicitly learns the light transport properties of the scene with the help of self-supervised SVBRDF components from our DecomposeNet. 

\begin{figure}[!h]
    \centering
    \includegraphics[width=0.9\linewidth]{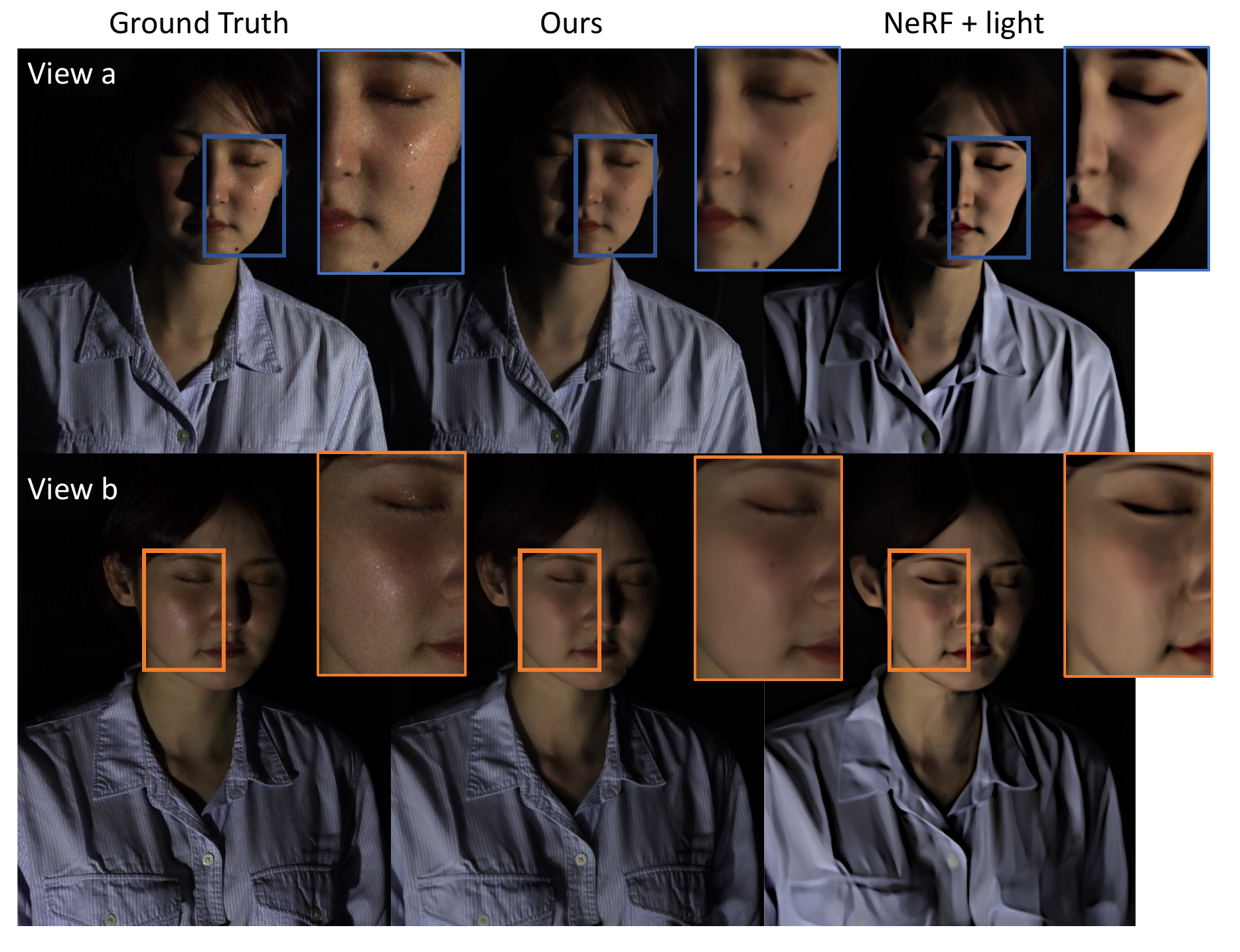}
    \caption{Qualitative comparison on simultaneous relighting and view synthesis. For each row, we showed a synthesis view with unseen lighting from ground truth, ours, and NeRF+light. Our results exhibit a more natural looking and are less blurry than NeRF+light.}
    \label{fig:CompareNERF}
\end{figure}
For recent NeRF-based methods such as \textcolor{black}{NeRFactor~\cite{zhang2021nerfactor}, InvRender~\cite{zhang2022modeling}, NeRD~\cite{boss2021nerd} and Neural-PIL~\cite{boss2021neural}, although they target for simultaneous novel view synthesis and relighting same as our approach, their inputs have fixed illumination which is different from our method. Therefore, for a fair comparison, we extend the volumetric rendering method NeRF~\cite{mildenhall2020nerf} with relighting capability, denoted as ``NeRF+light". 
To implement NeRF+light, we condition the NeRF's output radiance on the lighting direction. 
As shown in  Fig.~\ref{fig:CompareNERF}, we compare our method with NeRF+light on two novel views with novel light directions. 
It can be seen that our method produces more natural-looking and less blurry synthesis results than NeRF+light (\textit{e.g.}, the wrinkles on the cloth, and the facial details in the close-up insets). 
This is because our SVBRDF decomposition module enables the model to better understand the scene and maintain geometry/texture/material details under viewpoint/lighting variation.
}

\subsection{Speed Analysis}
\label{sec:speed}
\textcolor{black}{Our pipeline is based on the light field based neural networks~\cite{li2022neulf,sitzmann2021lfns}. Compared with the volumetric NeRF-based method, we directly regress ray color rather than aggregate radiance with time-consuming ray marching. 
Hence, as shown in Table.~\ref{tab:speed}, our method is significantly faster than NeRF+light, while still having higher rendering PSNR. }
% Admittedly, our method is limited synthesis front view only other than 360 degrees. }

\begin{table}[!h]
\centering
\caption{Comparison of rendering speed and quality with NeRF+light. We compute average PSNR across all real data. \label{tab:speed}}
\resizebox{\columnwidth}{!}{%
\begin{tabular}{l|cccc}
\hline
method  & Input & Viewpoint range &  second/frame  &  PSNR $\uparrow$   \\ \hline
Ours         & 4D + light      & front views       & 0.151      & 28.33          \\ %\hline
NeRF+light   & 5D + light      & $360^\circ$       & 22.143     & 26.12            \\ \hline

% NeuLF (Ours) & 8.1M       & 100ms   & 9.1M       & 100s     \\ \hline
\end{tabular}
}
\end{table}
 
\begin{figure}[!h]
    \centering
    \includegraphics[width=0.9\linewidth]{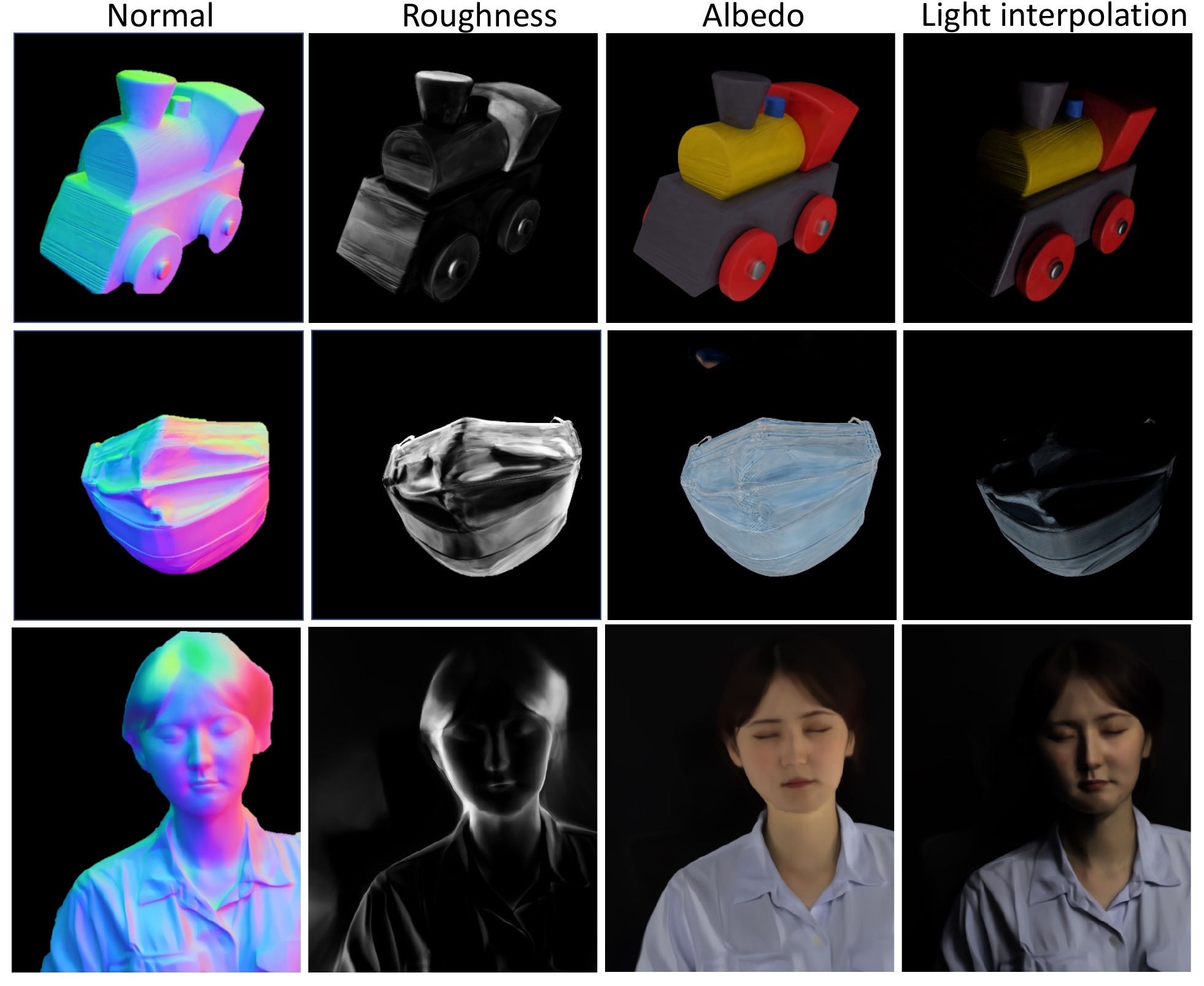}
    \caption{Qualitative results on SVBRDF decomposition. From left to right: surface normal map, surface roughness, surface albedo, and our rendering under a novel light direction.}
    \label{fig:ResDecompose}
\end{figure}

\subsection{SVBRDF Decomposition Results}
\label{sec:svbrdfd}
Fig.~\ref{fig:ResDecompose} shows the decomposition results on the synthetic toy train, face cover, and real human face scene. 
The results show that our DecomposeNet can generate reasonable view-independent normal, albedo, roughness field by using self-supervised microfacet rendering loss. 
We can see that in these three subjects, the estimated normal map is in line with the object geometry and contains shape details such as the wrinkles on the mask and cloth, and the bumpy surface on the wood train.
The estimated roughness map plausibly represents the roughness of the object's surface.
% restores the roughness of the surface of the object. 
And the estimated albedo restores the intrinsic color of the objects. 
Although the albedo still contains a small amount of baked-in lighting effects due to inter-reflection, it has little impact on our final rendering quality.
Rather, with such reasonable decomposition results, our RenderNet can already learn a good balance between the microfacet model and out-of-model effects.
The last column of Fig.~\ref{fig:ResDecompose} is the relighting result given a novel lighting direction. 
The results are realistic, where the shadows correctly depict the geometry and lighting.

\begin{figure}[!h]
    \centering
    \includegraphics[width=0.9\linewidth]{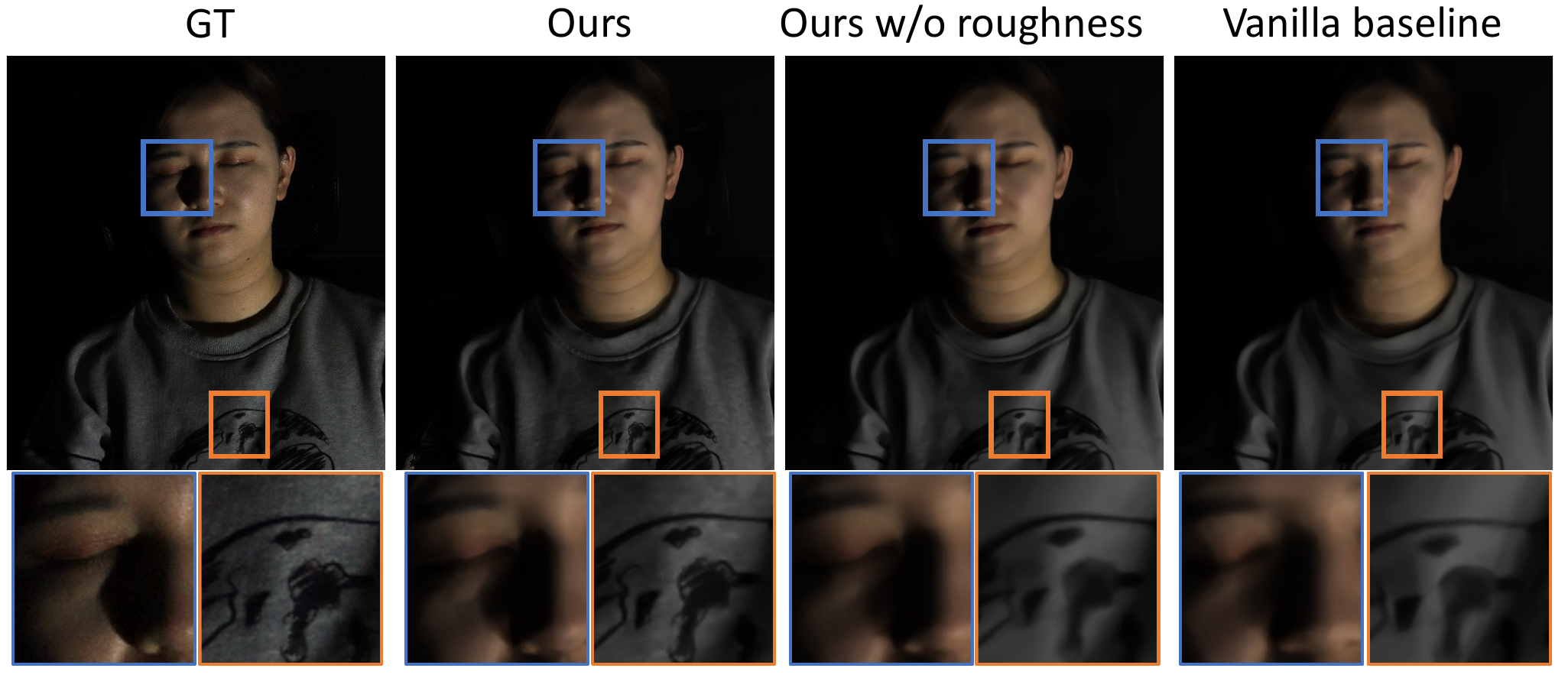}
    \caption{Ablation study on SVBRDF Decomposition. From left to right: ground truth, ours, ours w/o roughness, and our baseline. The bottom row shows close-up views for easier comparison.}
    \label{fig:Decom_ablation}
\end{figure}

~\textbf{Ablation study on SVBRDF decomposition} To demonstrate the importance of the SVBRDF decomposition from DecomposeNet, we conduct the following ablation study. We compared our full method with the vanilla baseline method described in Sec~\ref{Sec:Relit-NeuLF} (our baseline) and a baseline method that only decomposes normal and albedo without considering roughness (our w/o roughness).
This ablation is conducted on both real and synthetic datasets, where Table~\ref{tab:ablation_rgb} shows the average metrics.
Fig.~\ref{fig:Decom_ablation} shows the novel view relighting results on an example data. \textcolor{black}{In particular, compared with ground truth, we observe that ours w/o roughness and our baseline exhibit more over-smoothing effects than ours, such as the highlight details on the face and the pattern on the cloth.} 
Our full solution outperforms the other two baselines both quantitatively and qualitatively. This is because, with SVBRDF parameters as a prior, our RenderNet implicitly learns a more physically-correct rendering process.

\begin{table}[!h]
\small
\centering
\caption{Ablation study on SVBRDF Decomposition. The metrics are averaged over the test views and light directions in both real and synthetic datasets.}
\resizebox{0.7\columnwidth}{!}{%
\begin{tabular}{l|ccc}
\hline
% &       &    PSNR $\uparrow$     &       & & & SSIM $\uparrow$ &    && &LPIPS $\downarrow$  & \\ %\hline
method & {PSNR$\uparrow$} & {SSIM$\uparrow$} & {LPIPS$\downarrow$} \\
\hline
Ours baseline & 30.88 & 0.921 & 0.067 \\
Ours w/o roughness  & 31.15 & 0.917 & 0.042\\
Ours & \textbf{31.84} & \textbf{0.932} & \textbf{0.038} \\
\hline

\end{tabular}
}
\label{tab:ablation_rgb}
\end{table}

% \begin{figure}[!h]
%     \centering
%     \includegraphics[width=\linewidth]{Figure/failureCase.pdf}
%     \caption{Failure case when relighting with largely extrapolated light direction. The red arrows point to unnatural specular patterns.}
%     \label{fig:failcase}
% \end{figure}

% \subsection{Failure Cases}
% Although our proposed model has a strong ability to synthesize under novel interpolated or even extrapolated light directions, it is sometimes difficult to produce good results when the light direction is too far away from the training data distribution. Fig.~\ref{fig:failcase} shows three relighting results when the light comes from very low angles, which are unseen in training data.
% The appearance and shadow direction are generally plausible, but parts of the face (\textit{e.g.}, jaw, and eyelids) exhibit unnatural specular patterns and lack high-frequency details. 
% Nevertheless, it has little effect on relighting with the HDRI environment map.

\section{Conclusions}
In summary, we have presented Relit-NeuLF, a two-stage MLP network that simultaneously achieves highly-detailed relighting and novel view synthesis with weak supervision. Our method employs DecomposeNet to decompose SVBRDF components with self-supervision, based on sparse camera views and limited light directions. By utilizing the decomposed SVBRDF components, our implicit RenderNet generates high-fidelity novel view synthesis and relighting results, trained end-to-end with self-supervised rendering loss and photometric loss. Our approach achieves high rendering quality with fast rendering speed and low memory cost, thanks to the 4D light field representation. Although our method has limitations, such as requiring hard-to-acquire inputs and being unable to perform non-frontal view synthesis, we suggest future work to extend our method with a more flexible light field representation, explore light field network-based relighting with unconstrained input illumination, and investigate alternative spatial or frequency embedding approaches that generalize well in light field representation.

\section{Acknowledgement}
We wish to convey our gratitude to our previous intern, Yuqi Ding, for his foundational efforts on the dome structure and the development of the multi-lighting system.
\bibliographystyle{ACM-Reference-Format}
\bibliography{sample-base}

%%
%% If your work has an appendix, this is the place to put it.
\appendix

\end{document}